\documentclass[sigconf]{acmart}
\AtBeginDocument{%
  \providecommand\BibTeX{{%
    \normalfont B\kern-0.5em{\scshape i\kern-0.25em b}\kern-0.8em\TeX}}}

\setcopyright{acmcopyright}
\copyrightyear{2018}
\acmYear{2018}
\acmDOI{XXXXXXX.XXXXXXX}

\acmConference[Conference acronym 'XX]{Make sure to enter the correct
  conference title from your rights confirmation emai}{June 03--05,
  2018}{Woodstock, NY}
\acmPrice{15.00}
\acmISBN{978-1-4503-XXXX-X/18/06}

\usepackage{multirow}

\begin{document}

\copyrightyear{2023}
\acmYear{2023}
\setcopyright{acmlicensed}\acmConference[SIGIR '23]{Proceedings of the 46th
International ACM SIGIR Conference on Research and Development in
Information Retrieval}{July 23--27, 2023}{Taipei, Taiwan}
\acmBooktitle{Proceedings of the 46th International ACM SIGIR Conference on
Research and Development in Information Retrieval (SIGIR '23), July 23--27,
2023, Taipei, Taiwan}
\acmPrice{15.00}
\acmDOI{10.1145/3539618.3591809}
\acmISBN{978-1-4503-9408-6/23/07}

\title{
NeuralKG-ind: A Python Library for Inductive Knowledge Graph Representation Learning
}


\author{Wen Zhang}
\email{zhang.wen@zju.edu.cn}
\author{Zhen Yao}
\email{22151303@zju.edu.cn}
\affiliation{
\institution{Zhejiang University}
\country{China}
}

\author{Mingyang Chen}
\email{mingyangchen@zju.edu.cn}
\author{Zhiwei Huang}
\email{22251140@zju.edu.cn}
\affiliation{
\institution{Zhejiang University}
\country{China}
}

\author{Huajun Chen}
\email{huajunsir@zju.edu.cn}
\affiliation{
\institution{Zhejiang University}
\institution{Donghai Laboratory}
\institution{Alibaba-Zhejiang University Joint Institute of Frontier Technology}
\city{ }
\country{ }
}

\renewcommand{\shortauthors}{Trovato and Tobin, et al.}

\begin{abstract}
Since the dynamic characteristics of knowledge graphs, many inductive knowledge graph representation learning (KGRL) works have been proposed in recent years, focusing on enabling prediction over new entities. NeuralKG-ind is the first library of inductive KGRL as an important update of NeuralKG library. It includes standardized processes, rich existing methods, decoupled modules, and comprehensive evaluation metrics. With NeuralKG-ind, it is easy for researchers and engineers to reproduce, redevelop, and compare inductive KGRL methods. The library, experimental methodologies, and model re-implementing results of NeuralKG-ind are all publicly released at \href{https://github.com/zjukg/NeuralKG/tree/ind}{ https://github.com/zjukg/NeuralKG/tree/ind}. 
\end{abstract}





\maketitle

\section{Introduction}
Knowledge graphs (KGs) are formed as collections of triple representing facts in the world, denoted as \textit{(head entity, relation, tail entity)}. 
In the past decade, many large-scale KGs \cite{WordNet,conceptnet,freebase,nell} have been constructed and used in many applications \cite{LM-KG-1,LM-KG-2,ImgCap-KG-1, VQA-KG-1}, resulting in a lot of exciting research works related to KGs.  
One of the hot research topics is KG representation learning (KGRL), also called KG embedding (KGE), 
aiming to map elements in KGs into continuous vector spaces and provide knowledge service for tasks through vector space calculation. 
Typical methods include conventional KGEs \cite{TransE,RotatE,DistMult,ComplEx}, GNN-based KGEs \cite{RGCN,CompGCN}, and rule-based KGEs \cite{RUGE,IterE}. 
However, the world is dynamic, where new entities are continuously added to KGs, and new KGs are continuously constructed. 
The traditional KGRL methods, which learn embeddings for a fixed set of entities, fail to generalize to new elements.
Thus recent years have witnessed rapid growth in exploring inductive KGRL \cite{knowledge-ext-survey,GraIL-ICML2020,CoMPILE-AAAI2021,MorsE-SIGIR2022}, making KGRL methods extrapolate to new entities. 
Though many frameworks \cite{mukg,arm2020torchkge,ampligraph,DGL-KE,graphvite}, including OpenKE \cite{han2018openke}, Pykg2vec \cite{yu2019pykg2vec}, LibKGE \cite{libkge}, PyKEEN \cite{ali2021pykeen}, and NeuralKG \cite{NeuralKG}, unify the programming process of traditional KGRL methods, considering the process of inductive KGRL is related to but also significantly different from traditional KGRL methods, a toolkit specifically for inductive KGRL is highly expected.
Thus we develop the first inductive KGRL toolkit NeuralKG-ind, which is a version update based on NeuralKG with the following features:

\begin{figure*}[t]
    \centering
    \vspace{-1mm}
\includegraphics[scale=0.33]{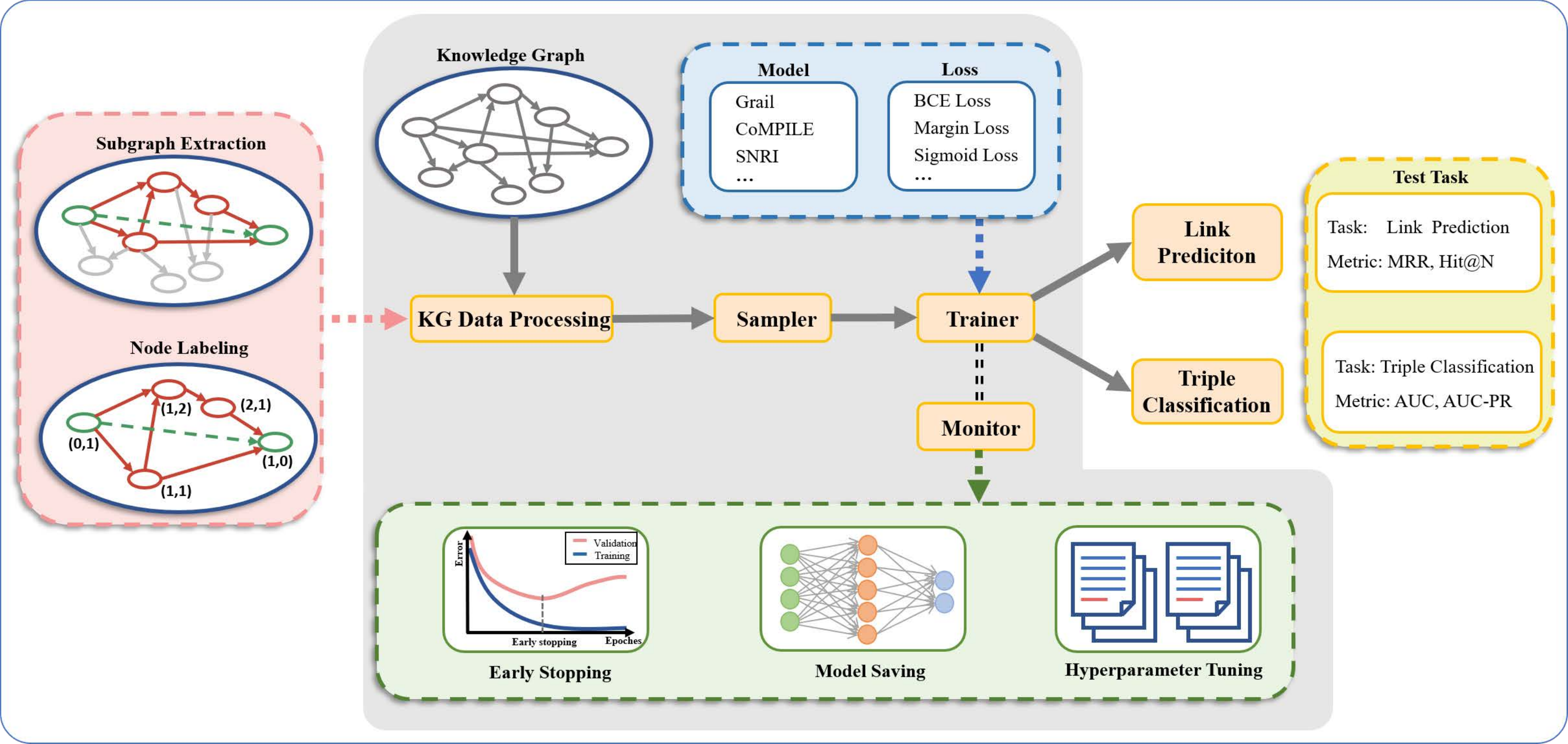}
	\caption{The pipeline of NeuralKG-ind retains the original NeuralKG framework with the gray background while adapting it to the inductive KGRL method. The non-gray background is unique to NeuralKG-ind.}
 \vspace{-3mm}
	\label{fig:overview}
\end{figure*}

\begin{itemize}
    \item \textbf{Standardized process.} According to existing methods, we standardized the overall process of constructing an inductive KGRL model, including data processing, sampler and trainer construction, and evaluation of link prediction and triple classification tasks. 
    We also provide auxiliary functions for model training and analysis, including log management and hyper-parameter tuning.
    \item \textbf{Rich existing methods.} We re-implemented 5 inductive KGRL methods proposed in recent 3 years, including GraIL \cite{GraIL-ICML2020}, CoMPILE \cite{CoMPILE-AAAI2021}, SNRI \cite{SNRI-IJCAI2022}, RMPI \cite{RMPI-ICDE2023}, and MorsE \cite{MorsE-SIGIR2022}, enabling users to apply these models off the shelf. 
    \item \textbf{Decoupled modules.} We provide a lot of decoupled modules, such as the subgraph extraction function, the node labeling function, neighbor aggregation functions, compound graph neural network layers, and KGE score functions, enabling users to construct a new inductive KGRL model faster.
    \item \textbf{Comprehensive evaluations.} We conduct comprehensive evaluations with NeuralKG-ind and report diverse results, including ACU, MRR, Hit@1, and Hit@10, presenting more precise pictures of different methods.
    \item \textbf{Long-term supports.} We provide long-term support on NeuralKG-ind, including maintaining detailed documentation, creating straightforward quick-start, adding new models, solving issues, and dealing with pull requests.
\end{itemize}

\section{Preliminary}
A knowledge graph $\mathcal{G} = \{\mathcal{E}, \mathcal{R}, \mathcal{T}\}$ consists of an entity set $\mathcal{E}$ and a relation set $\mathcal{R}$, and the triples are formulated as $\mathcal{T} = \{(h,r,t)\} \subseteq \mathcal{E} \times \mathcal{R} \times \mathcal{E}$. 
In \textit{inductive} settings \cite{GraIL-ICML2020,MorsE-SIGIR2022}, there is a training KG $\mathcal{G}^{tr} = \{\mathcal{E}^{tr}, \mathcal{R}^{tr}, \mathcal{T}^{tr}\}$ and a set of test triples $\mathcal{G}^{te} = \{\mathcal{E}^{te}, \mathcal{R}^{te}, \mathcal{T}^{te}\}$, where
entities in $\mathcal{G}^{te}$ are not seen during training, i.e., $\mathcal{E}^{tr} \cap \mathcal{E}^{te} = \emptyset$. Furthermore, to provide connections between unseen entities and specify links for unseen entities to be predicted, the test triples $\mathcal{T}^{te}$ are usually split into two parts, 
 $\mathcal{T}^{sup}$ for supporting and $\mathcal{T}^{que}$ for querying.
Models are trained on the training KG $\mathcal{G}^{tr}$,
and then adapted to the test KG $\mathcal{G}^{te} = \{\mathcal{E}^{te}, \mathcal{R}^{te}, \mathcal{T}^{sup}, \mathcal{T}^{que}\}$.
We generally divided inductive KGRL methods into
\begin{itemize}
    \item \textbf{Subgraph predicting methods.} 
    For predicting the relation between two unseen entities, subgraph predicting methods \cite{GraIL-ICML2020,CoMPILE-AAAI2021,SNRI-IJCAI2022,RMPI-ICDE2023,TACT-AAAI2021,ConGLR-SIGIR2022} first extract a subgraph around them, then label each entity in this subgraph through manually designed features and finally use a GNN to encode and score the subgraph with candidate relations.
    \item \textbf{Entity encoding methods.} These methods aim at learning transferable functions that can encode entity embeddings but not learn fixed entity embedding tables. Existing methods \cite{MorsE-SIGIR2022,MaKEr-IJCAI2022} sample subgraphs from the training KG and form each subgraph as a task containing support and query triples. Then they use meta-learning training paradigms to learn transferable information across tasks.
\end{itemize}

\begin{figure*}[t]
    \centering
\includegraphics[scale=0.32]{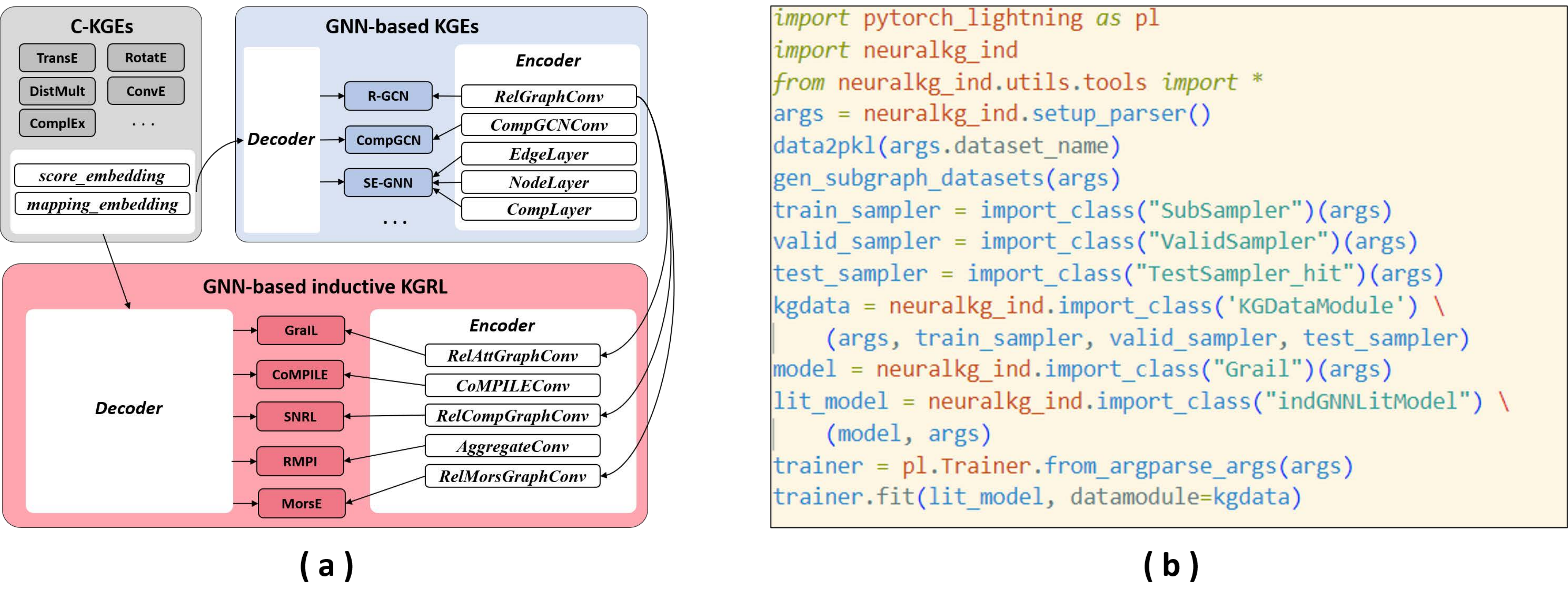}
\vspace{-5mm}
	\caption{(a) Overview of relational graph neural network framework in NeuralKG-ind.  (b) Quick start example.}
	\label{fig:start}
 \vspace{-3mm}
\end{figure*}

\section{Structure of NeuralKG-ind}

\subsection{NeuralKG-ind Pipeline}
NeuralKG-ind provides a general workflow for developing models handling inductive tasks on KGs. Figure \ref{fig:overview} shows the pipeline. 
We develop NeuralKG-ind based on the NeuralKG \cite{NeuralKG} framework, where KGRL models are standardized with KG data processing, sampler, trainer, and monitor. We also follow this process in NeuralKG-ind. Next, we introduce the modules and functions in detail.

\subsubsection{KG Data Processing}
The KG data processing module takes the raw KG data as input and performs necessary preprocessing, and compared to NeuralKG, we add graph processing modules in NeuralKG-ind, including indexing elements of a KG and packaging it into a python-friendly format (i.e., pickle) (\textit{Data2pkl}), constructing input for various models (\textit{Gen\_sub\_datasets}, \textit{Node\_label}), and wrapping them into elaborate python classes for efficient training (\textit{GRData}, \textit{BaseGraph} and \textit{KGDataModule}). 

\paragraph{Data2pkl}
This function creates an element mapping dictionary to assign a unique numerical id, usually starting from 0, for each entity and relation in the dataset. 
Then the triples are represented by ids based on the mapping dictionary.
Finally, the indexed triples and mapping dictionary are serialized into pickle format to improve the efficiency of following operations on datasets.

\paragraph{Gen\_subgraph\_datasets}
This function extracts subgraphs for triples in a KG and stores them in a Lightning Memory-Mapped Database (LMDB). 
For subgraph predicting methods, this function generates the enclosing subgraph, which consists of all paths from the $h$ to $t$ of a specific triple $(h, r, t)$.
For entity encoding methods, this function generates random subgraphs and splits each of them into support and query triples to form a task for meta-learning.

\paragraph{Node\_label}
This function is mainly for subgraph predicting methods and is used to convert the topological information of an attribute-free graph into a node feature representation: 
\begin{equation}
h_{i} = [\text { one-hot }(d(i, h)) \oplus \text { one-hot }(d(i, t))],
\end{equation}
where $\oplus$ denotes concatenation of two vectors, $d(\cdot,\cdot)$ denotes the shortest distance between two nodes. It is noted that not all inductive models require node feature representation.

\paragraph{GRData}
\textit{GRData} is named after \underline{Gr}aph \underline{Data}. It loads the encoded triples and the subgraph database, and provides a function to perform several rounds of random negative sampling on the training set and generate corresponding subgraphs for training. It also provides statistical information about the subgraphs, such as the maximum and average subgraph size, pruning ratio, and so on.

\paragraph{BaseGraph}
This class is one of the core classes for KG data processing. It generates triple datasets for the training, validation, and testing processes and performs batch loading in the \textit{KGDataModule}. The \textit{GRData} class generates the corresponding positive and negative sample subgraphs and triple relations in the triple classification task. In the link prediction task, it performs negative sampling on the head or tail entity of the test triple and generates subgraphs as the input to the model.

\paragraph{KGDataModule}
This class is inherited from the \textit{LightningDataModule} in PyTorch Lightning\footnote{https://www.pytorchlightning.ai/}, which standardizes the training, validation, test data preparation and loading. It integrates the functionalities of \textit{Dataset} and \textit{Dataloader} in PyTorch \cite{pytorch}. It supports a customizable function for preparing the data into mini-batches for the training, validating, and testing phases in \textit{Trainer} module.

\begin{table*}[t]
  \centering
  \vspace{-1mm}
  \resizebox{0.9\textwidth}{!}{
  \begin{tabular}{lcccccccc}
    \toprule
    \multirow{2}{*}{\textbf{Method}} & \multicolumn{2}{c}{\textbf{FB15k-237-v1}} & \multicolumn{2}{c}{\textbf{FB15k-237-v2}} & \multicolumn{2}{c}{\textbf{FB15k-237-v3}} & \multicolumn{2}{c}{\textbf{FB15k-237-v4}}  \\
    \cmidrule(lr){2-3} \cmidrule(lr){4-5} \cmidrule(lr){6-7} \cmidrule(lr){8-9}
     & AUC-PR & Hit@10 & AUC-PR & Hit@10 & AUC-PR & Hit@10 & AUC-PR & Hit@10 \\ 
    \midrule 
    \text {GraIL} & 0.821 | 0.847 & 0.624 | 0.642 & 0.900 | 0.906 & 0.831 | 0.818 & 0.899 | 0.917  & 0.828 | 0.828 & 0.921 | 0.945 & 0.880 | 0.893   \\
    \text {CoMPILE} & 0.835 | 0.855 & 0.668 | 0.676 & 0.905 | 0.917 & 0.813 | 0.830 & 0.925 | 0.931  & 0.859 | 0.847 & 0.932 | 0.949 & 0.894 | 0.874   \\
    \text {SNRI} & 0.833 | 0.867 & 0.720 | 0.718 & 0.906 | 0.918 & 0.857 | 0.865 & 0.884 | 0.912  & 0.871 | 0.860 & 0.916 | 0.933 & 0.894 | 0.894   \\
    \text {RMPI} & 0.823 | 0.859 & 0.689 | 0.717 & 0.882 | 0.929 & 0.830 | 0.837 & 0.866 | 0.927  & 0.827 | 0.860 & 0.916 | 0.933 & 0.866 | 0.886   \\
    \text {MorsE} & 0.847 |  \quad $\backslash$ \ \quad    & 0.833 | 0.847 & 0.960 | \quad $\backslash$ \ \quad & 0.950 | 0.963 & 0.952 | \quad $\backslash$ \ \quad & 0.954 | 0.957 & 0.952 | \quad $\backslash$ \ \quad & 0.958 | 0.964   \\
    \midrule 
  \end{tabular}}
  \caption{AUC-PR and Hit@10 results on the inductive benchmark datasets extracted from FB15k-237. The left results are from NeuralKG-ind, and the right results are from paper.}
  \vspace{-5mm}
  \label{tb:main result}
\end{table*}

\begin{table}[t]
  \centering
  \resizebox{0.48\textwidth}{!}{
  \begin{tabular}{c lcccccc}
    \toprule
     \textbf{Dataset} & \textbf{Method} & \textbf{AUC} & \textbf{AUC-PR} & \textbf{MRR} & \textbf{Hit@1} & \textbf{Hit@5} & \textbf{Hit@10} \\ 
    \midrule 
    \multirow{2}{*}{\textbf{v1}} &\text{GraIL} & 0.814 & 0.750 & 0.467 & 0.395 & 0.515 & 0.575  \\
    & \text{SNRI} & 0.737 & 0.720 & 0.523 & 0.475 & 0.545 & 0.595  \\
    \hline
     \multirow{2}{*}{\textbf{v2}}  &\text{GraIL} & 0.929 & 0.947 & 0.735 & 0.624 & 0.884 & 0.933  \\
     & \text{SNRI} & 0.864 & 0.884 & 0.630 & 0.507 & 0.774 & 0.863 \\
    \midrule 
  \end{tabular}}
  \caption{Results with comprehensive evaluation metrics on the inductive benchmark datasets extracted from NELL-995.}
  \vspace{-5mm}
  \label{tb: more result}
\end{table}

\subsubsection{Sampler}
The sampler module is called with each mini-batch sample in the data loading process. We expected to collect the mini-batch samples into samples with subgraphs and negative sampling. We provide the unified sampler module applied in various inductive models. The sampler module includes \textit{SubSampler}, \textit{ValidSampler}, \textit{TestSampler\_hit}, and \textit{TestSampler\_auc}. In addition, users could also define customized sampler modules to achieve custom batching.

\paragraph{SubSampler/ValidSampler}
\textit{SubSampler} class is inherited from the \textit{Basegraph} class. Before training, \textit{SubSampler} generates the positive and negative subgraph samples and triple relations and stores them in the dictionary. \textit{ValidSampler} class is for dataset sampling in the validation phase. It generates and negatively samples the subgraphs and relations of the validation triples and generates global positive and negative labels for binary classification validation. In the meta-learning setting, it will degenerate into an identity function.

\paragraph{TestSampler\_hit / TestSampler\_auc}
\textit{TestSampler} class is divided into two categories. Based on different testing tasks, we have different testing data and metrics. \textit{TestSampler\_hit} is used for data sampling in link prediction tasks, with metrics such as MRR and Hit@N. During testing, it will perform multiple random negative samplings (50 times by default) on the test triple's head entity and tail entity, respectively, and generate subgraphs. In addition, it will also record the index of the ground truth. \textit{TestSampler\_auc} is used for data sampling in triple classification tasks, with metrics such as AUC and AUC-PR. Its sampling process is similar to \textit{ValidSampler}.

\subsubsection{Trainer}
The trainer module calls the \textit{LightningModule} to guide the training and validation. It automatically places the training data and model on the GPU. It provides real-time feedback on the progress facilitating users to quickly experiment and comprehensively compare different types of inductive models.

\paragraph{Training}
The trainer module is inherited from the \textit{Lightning} \textit{Module} class in PyTorch Lightning, which sets the model's training process and realizes automatic back-propagation gradient update. The model training information is displayed on the progress bar of models in real-time. Users can save the training logs locally or upload them to the cloud for subsequent experimental analysis.

\paragraph{Validation}
The trainer module accepts external parameters such as \textit{check\_per\_epoch} or \textit{check\_per\_step} to set the number of rounds of model evaluation.
In the inductive task setting, the triple classification task is used as the default method for evaluating the model's performance by default. The area under the curve (AUC) and the area under the precision-recall curve (AUC-PR) metrics are applied to evaluate the triple classification results of models.

\subsubsection{Evaluation Tasks}
In NeuralKG-ind, we add the triple classification task and the link prediction base inductive task setting to evaluate the performance of the model. 

\paragraph{Link Prediction Task}

Predicting the missing links between entities is a fundamental task for knowledge graphs called link prediction. In the inductive task setting, \textit{link\_predict} function performs multiple random negative samplings as the candidate's head or tail entities, and \textit{ind\_predict} function generates the ranking of ground truth. Finally, \textit{get\_result} function calculates the MRR and Hit@N to evaluate the model's performance. We have provided a demo on the GitHub page to show how the model predicts the unseen entities through the training and supporting triples.

\paragraph{Triple Classification Task}

The triple classification task evaluates the model's ability to distinguish between positive and negative samples. \textit{classification} function calculates the scores for samples processed by \textit{ValidSampler} class. \textit{get\_auc} function compares the scores and gets the binary classification results AUC and AUC-PR.

\subsubsection{Monitor}
The monitor module continuously records the evaluation results of the model and decides whether to stop the training, the same as in NeuralKG.

\subsection{Relational GNN framework}
Most inductive KGRL utilizes \underline{r}elational \underline{GNN} (RGNN) as the subgraph encoder. RGNN is also used in GNN-based KGEs.
Thus, we summarized a construction paradigm for RGNN and rebuilt GNN-based KGEs in NeuralKG. It includes model initialization, building hidden layers, building a multi-layer graph model, and the model forward propagation. 
This paradigm makes it easier for users to develop new RGNNs quickly for GNN-based KGE and inductive KGRL methods.
Figure \ref{fig:start} (a) shows the overview.

\subsubsection{GNN-based KGE models}
This model hub covers the traditional graph representation learning model, including the R-GCN \cite{RGCN}, CompGCN \cite{CompGCN} and SE-GNN \cite{SE-GNN_AAAI2022}. R-GCN and CompGCN are models refactored based on NeuralKG. SE-GNN is a new model updated from NeuralKG. In NeuralKG-ind, the structure of the neural network layers has been decoupled. The base layer of R-GCN, \textit{RelGraphConv}, has been decoupled as a separate class. Since NeuralKG-ind is developed based on NeuralKG, users can directly set C-KGEs in NeuralKG as decoders without reinventing the wheel. For example, in CompGCN, the decoder directly calls the ConvE \cite{ConvE} and DistMult \cite{DistMult} models of the C-KGE module.

\subsubsection{GNN-based inductive KGRL models}
This model hub covers the inductive models, including GraIL \cite{GraIL-ICML2020}, CoMPILE \cite{CoMPILE-AAAI2021}, SNRI \cite{SNRI-IJCAI2022}, RMPI \cite{RMPI-ICDE2023} and MorsE \cite{MorsE-SIGIR2022}.
The GNN structure has been proven to be effective in inductive task settings. Thus, a series of works tried implementing GNN as encoders for element representations and applying conventional KGEs as decoders for triple scoring. 
The GNN-based inductive KGRL model class has two main advantages: highly decoupled modules, directly calling conventional KGE model. Using \textit{RelGraphConv} as the parent class, we propose several incremental layers: \textit{RelAttGraphConv} with edge attention mechanisms and \textit{RelCompGraphConv} with composition operations. They form the incremental R-GCN model as the base layer for GraIL and SNRI, respectively. Moreover, the conventional KGE models in NeuralKG, such as TransE \cite{TransE} and RotatE \cite{RotatE}, are directly called as decoders. Researchers can conveniently call different decoders to understand the inductive models' performance better.

\subsection{Evaluation}
With a unified model library, NeuralKG supports users to run existing models directly or build customized pipelines quickly. An example is shown in Figure \ref{fig:start} (b). We also provide detailed documentation in Github page and a quick-start example in colab to help users use NeuralKG-ind. As shown in Table \ref{tb:main result}, NeuralKG provides the implementation of the inductive models and the evaluation in inductive task setting on FB15k-237 series datasets, comparing their performance with the reported results in the paper. Compared to reported results in original papers, NeuralKG successfully replicated similar results.
NeuralKG provides multiple datasets and comprehensive evaluation metrics, which facilitates researchers to understand better the model's performance in the inductive task setting. Table \ref{tb: more result} shows more comprehensive GraIL and SNRI evaluation results on the NELL-995 datasets. Compared with the results reported in the paper, it can be seen that the accuracy of the models is still not high in the more strict MRR and Hit@1 metrics. The performance of the models varies depending on the dataset. For example, SNRI performs better on the FB15k-237 dataset, while GraIL performs better on the NELL-995 dataset.

\section{Conclusion}
In this work, we introduce the first library for inductive KGRL, NeuralKG-ind, as an update of NeuralKG. With NeuralKG-ind, users could implement existing models off the shelf, develop new models quickly with decoupled modules, and test their models with standard evaluation tasks and metrics. 
Besides the methods mentioned in the current version of our framework, some recently proposed models also focus on scenarios where unseen entities are connected to seen entities \cite{KnowTransOOKB-IJCAI2017,LAN-AAAI2019,CFAG-AAAI2022}, unseen entities have textual information \cite{StAR-WWW2021,BLP-WWW2021,SimKGC-ACL2022}, or unseen relations emerge \cite{GraphANGEL-ICLR2022,CSR-NIPS2022}. 
Thus in the future, we plan to include these new types of inductive KGRL methods into NeuralKG-ind following the current framework.

\begin{acks}
This work was supported by the National Natural Science Foundation of China (NSFCU19B2027, NSFC91846204), joint project DH-2022ZY0012 from Donghai Lab, Zhejiang Provincial Natural Science Foundation of China (No. LQ23F020017) and Yongjiang Talent Introduction Programme.
\end{acks}

\clearpage
\bibliographystyle{ACM-Reference-Format}
\bibliography{sample-base}

\end{document}